\documentclass[runningheads]{llncs}
\usepackage[T1]{fontenc}

\usepackage{graphicx}
\graphicspath{ {images/} }

\usepackage{amsmath} 
\usepackage{color}
\usepackage{booktabs}
\usepackage{hang}
\usepackage{subcaption}
\usepackage{pifont}

\usepackage{xcolor} 
\usepackage[table]{xcolor} 
\usepackage{verbatim}
\usepackage{siunitx}

\usepackage{algorithm}
\usepackage{algpseudocode}
\usepackage{url}
\usepackage{hyperref}

\begin{document}

\title{Discriminator-Guided Adaptive Diffusion for \\ Source-Free Test-Time Adaptation \\ under Image Corruptions}

\titlerunning{Adaptive Diffusion for Source-Free Test-Time Adaptation}

\author{Francesco~Olivato\inst{1}\orcidID{0009-0005-5674-7212} \and
Cigdem~Beyan\inst{1}\orcidID{0000-0002-9583-0087} \and
Vittorio~Murino\inst{1,2}\orcidID{0000-0002-8645-2328}}

\authorrunning{F. Olivato et al.}

\institute{Department of Computer Science, University of Verona, Verona, Italy \and 
AI for Good (AIGO), Istituto Italiano di Tecnologia, Genova, Italy
\\
\email{\{francesco.olivato, cigdem.beyan, vittorio.murino\}@univr.it}}

\maketitle  

\begin{abstract}
In this work, we study Source-Free Unsupervised Domain Adaptation under corruption-induced domain shifts, where performance degradation is caused by natural image corruptions that go beyond additive noise, including blur, weather effects, and digital artifacts. We propose a diffusion-based, input-level adaptation framework that operates entirely at test time and keeps all source-trained models frozen, explicitly targeting robustness to corrupted target inputs.
Our method leverages a source-trained diffusion model as a generative prior and introduces a discriminator-guided adaptive diffusion strategy that dynamically controls the amount of perturbation applied to each test sample. Rather than relying on a fixed diffusion depth, the discriminator determines, on a per-image basis, when sufficient forward diffusion has been applied to suppress corruption-specific artifacts, with each corruption type effectively defining a distinct target domain. This adaptive stopping mechanism applies only the necessary amount of noise to remove domain-specific corruption while preserving class-discriminative structure. The reverse diffusion process then reconstructs a source-aligned image, optionally stabilized through structural guidance, which is classified using a frozen source-trained classifier.
We evaluate the proposed approach across a broad spectrum of corruption-induced target domains, covering 15 diverse corruption types, and demonstrate more balanced robustness with competitive or improved performance across non-noise corruptions.
Additional analyses reveal how the adaptive diffusion schedule responds to different corruption characteristics, highlighting the practicality, generality, and robustness of the proposed framework. The code is publicly available at \url{https://github.com/fmolivato/dgadiffusion/}.


\end{abstract}

\section{Introduction}
\label{sec:intro}

In real-world deployment, deep neural networks are frequently exposed to \textbf{natural image corruptions}, such as changes in imaging conditions, blur, sensor noise, or environmental effects. Even when semantic content remains unchanged, these corruptions can cause severe performance degradation, as the resulting inputs differ substantially from the clean data seen during training.
At the same time, access to source-domain data is often restricted due to privacy, licensing, or storage constraints, motivating the setting of Source-Free Unsupervised Domain Adaptation (SFUDA), where a pretrained model must be adapted to an unlabeled corruption-shifted target domain without revisiting the source data~\cite{hendrycks2019benchmarking,Eastwood2021SourceFree,Hendrycks2021ManyFaces,Iwasawa2021TTAM,Kundu2020USFDA}.

A large class of SFUDA methods addresses this problem by adapting model parameters using unlabeled target data, for example, through entropy minimization~\cite{hendrycks2019benchmarking} or pseudo-label refinement~\cite{huai2023context,pei2023uncertainty}. While effective in certain settings, particularly for structured or semantic domain shifts, parameter adaptation introduces risks of catastrophic forgetting, requires careful hyperparameter tuning, and is often sensitive to target-domain characteristics, limiting scalability across diverse corruption types or continually shifting target distributions~\cite{gao2022back,Varsavsky2020TTUDA}.
An alternative paradigm focuses on input-level adaptation, where target samples are transformed to better match the source distribution while retaining class-discriminative structure, allowing the source-trained classifier to operate without modification~\cite{hoffman2018cycada,Rusak2020AugMix}.

Diffusion models~\cite{gao2022back,peng2024unsupervised} provide a natural framework for input-level adaptation under \textbf{corruption-induced domain shifts}, as their iterative noising and denoising processes can suppress corruption- and distribution-specific artifacts and reconstruct semantically consistent images aligned with learned source statistics. However, existing diffusion-based SFUDA approaches typically rely on fixed noising schedules, creating an inherent trade-off: insufficient noising fails to remove corruption-specific degradation, while excessive noising erases class-discriminative information~\cite{gao2022back,nie2022diffusion}, particularly when corruption types and severities vary across target samples.

In this work, we address this limitation by noting that the optimal amount of diffusion noise is \textbf{inherently sample-dependent} under corruption-induced shifts. We introduce \textbf{discriminator-guided adaptive diffusion scheduling}, a \textbf{test-time input-level adaptation strategy} that dynamically determines the appropriate noising depth for each target sample. During the forward diffusion process, a discriminator monitors residual domain-specific cues and adaptively halts noising once these cues are sufficiently suppressed. The subsequent reverse diffusion reconstructs a \textbf{source-aligned image} while minimizing semantic distortion, and can optionally incorporate a \textbf{structural guidance mechanism} to stabilize reconstruction under severe corruptions. By applying only the necessary amount of perturbation on a per-sample basis, our approach preserves class-discriminative structure under mild corruptions while enabling stronger adaptation for severe degradations, all while keeping both the diffusion model and the classifier frozen.

Our experimental evaluation focuses on robustness under corruption-induced domain shifts, comparing input-level adaptation with model-based alternatives in the source-free test-time setting. Beyond accuracy improvements, our approach offers practical advantages in adaptivity and deployment, as it avoids model updates and does not rely on fixed or manually tuned diffusion schedules. By determining the appropriate amount of adaptation on a per-sample basis, our method naturally accommodates varying types and severities of corruption. As a result, it achieves consistently balanced robustness across diverse natural image corruptions, highlighting its practicality for real-world test-time deployment.

The main contributions of this paper are:
\begin{itemize}
    \item We propose discriminator-guided adaptive diffusion scheduling for source-free test-time adaptation, enabling per-sample control of diffusion depth under natural image corruptions.
    \item We introduce an input-level adaptation framework that keeps both the diffusion model and the classifier frozen, avoiding parameter updates while balancing domain alignment and semantic preservation through structural guidance.
    \item We demonstrate improved and more balanced robustness across diverse corruption types through extensive experiments on 15 corruption types, supported by analyses of the adaptive stopping behavior.
\end{itemize}

%%%%%%%%%%%%%%%%%%%%%%%%%%%%%%%%%%%%%%%%%%%%%%%%%%%%%%%%%%%%%%%%

\section{Related Work}
\label{sec:relatedWork}

\subsubsection{Source-Free Unsupervised Domain Adaptation.}
SFUDA addresses the problem of adapting a pretrained source model to an unlabeled target domain without access to source data at deployment time~\cite{liang2020we,fang2024source,hendrycks2019benchmarking}. This setting is motivated by practical constraints such as data privacy, storage limitations, and regulatory requirements that prevent retaining or redistributing source data once a model is deployed. The central challenge in SFUDA is to align a source-trained model with a target distribution that may exhibit significant domain shift, without revisiting the original training data. Such shifts can arise from diverse factors, including changes in data acquisition conditions, sensor characteristics, environmental variations, semantic composition, or dataset bias. Herein, we focus specifically on \emph{corruption-induced domain shifts}, where the semantic content remains unchanged, but target inputs are degraded by natural image corruptions \cite{hendrycks2019benchmarking,Schneider2020CovShift}.

A predominant strategy in SFUDA involves \textbf{adapting model parameters} directly on target data, either in a dedicated adaptation phase or dynamically at test time~\cite{Iwasawa2021TTAM,Li2020ModelAdapt,Niu2022ETTA,Niu2023StableTTA,Pandey2021LabelPreserve}. Test-time adaptation (TTA) methods, for example, update model parameters on-the-fly using incoming target samples. A widely used principle in this area is entropy minimization, where the model is encouraged to produce confident predictions on target data~\cite{wang2021tent}. Tent~\cite{wang2021tent} exemplifies this approach by adapting batch normalization parameters to minimize prediction entropy, though it can be sensitive to batch size and data ordering. Other approaches extend model adaptation through pseudo-label refinement~\cite{huai2023context}, uncertainty-driven objectives~\cite{pei2023uncertainty}, or augmentation-based adaptation~\cite{zhang2021memo}.

While effective in some scenarios, model adaptation methods often require careful hyperparameter tuning, are sensitive to target-domain characteristics, and are prone to catastrophic forgetting, where the model’s performance on the original source domain deteriorates as its parameters drift away from the source solution~\cite{gao2022back,Eastwood2021SourceFree}. Additionally, many such methods are computationally demanding or rely on domain-specific optimization, limiting their scalability across diverse or continually shifting target domains. In contrast, our work diverges from this paradigm by keeping the source-trained classifier entirely fixed, thereby inherently avoiding catastrophic forgetting while enabling robust adaptation through input-level transformation.

\vspace{-1em}
\subsubsection{Input-Level Adaptation.}
An alternative direction within SFUDA focuses on adapting the input data rather than the model. In this paradigm, target-domain samples are transformed to resemble the source distribution, allowing a frozen source-trained classifier to operate without modification. Early approaches explored image translation and style transfer using generative adversarial networks, such as CycleGAN-based methods for cross-domain adaptation~\cite{hoffman2018cycada,Rusak2020AugMix}. However, these methods typically require joint access to source and target data during training and may not generalize well to unseen target distributions. More recently, diffusion-based input purification methods such as DiffPure~\cite{nie2022diffusion} have demonstrated strong robustness by reconstructing clean images through generative denoising. While effective for adversarial perturbations, DiffPure relies on a fixed denoising strategy that lacks adaptivity to the severity or type of corruption, limiting its effectiveness under diverse natural corruptions.

Our approach follows the input-adaptation paradigm but departs from prior work by introducing discriminator-guided adaptive diffusion. Instead of applying a fixed number of denoising steps, we dynamically determine the minimal amount of perturbation required on a per-sample basis to suppress domain-specific artifacts. This adaptive strategy enables effective correction of natural corruptions while preserving class-discriminative structure, all while keeping the source-trained classifier entirely frozen.

\vspace{-1em}
\subsubsection{Diffusion Models for SFUDA under Image Corruptions.}
Denoising Diffusion Probabilistic Models (DDPMs) provide a principled generative framework based on iterative noising and denoising, making them well-suited for transforming corrupted inputs toward a learned data distribution~\cite{ho2020denoising,saharia2022image}. The work most closely related to ours is Diffusion-Driven Adaptation (DDA)~\cite{gao2022back}, which demonstrated that a source-trained diffusion model can act as a powerful source prior for correcting common image corruptions. DDA applies a fixed number of forward diffusion steps followed by reverse denoising to project target images back toward the source domain. A key limitation of DDA is its reliance on a manually selected, fixed diffusion depth. Insufficient noising fails to eliminate domain-specific artifacts, while excessive noising risks destroying class-discriminative information~\cite{gao2022back}. To mitigate this sensitivity, DDA incorporates classifier guidance and self-ensembling to recover lost semantic cues.

Our approach builds on this foundation, but directly addresses the scheduling limitation by introducing discriminator-guided adaptive diffusion. Instead of using a fixed diffusion depth, we train a discriminator to distinguish noised target samples from noised source-like samples and halt the forward diffusion process on a per-sample basis once domain-specific cues are sufficiently suppressed. This adaptive, data-driven scheduling strategy removes the need for manual tuning and enables a more balanced trade-off between domain alignment and semantic preservation, while keeping both the diffusion model and the classifier frozen during adaptation.

\begin{figure}[t!]
    \centering
    \includegraphics[width=0.87\linewidth]{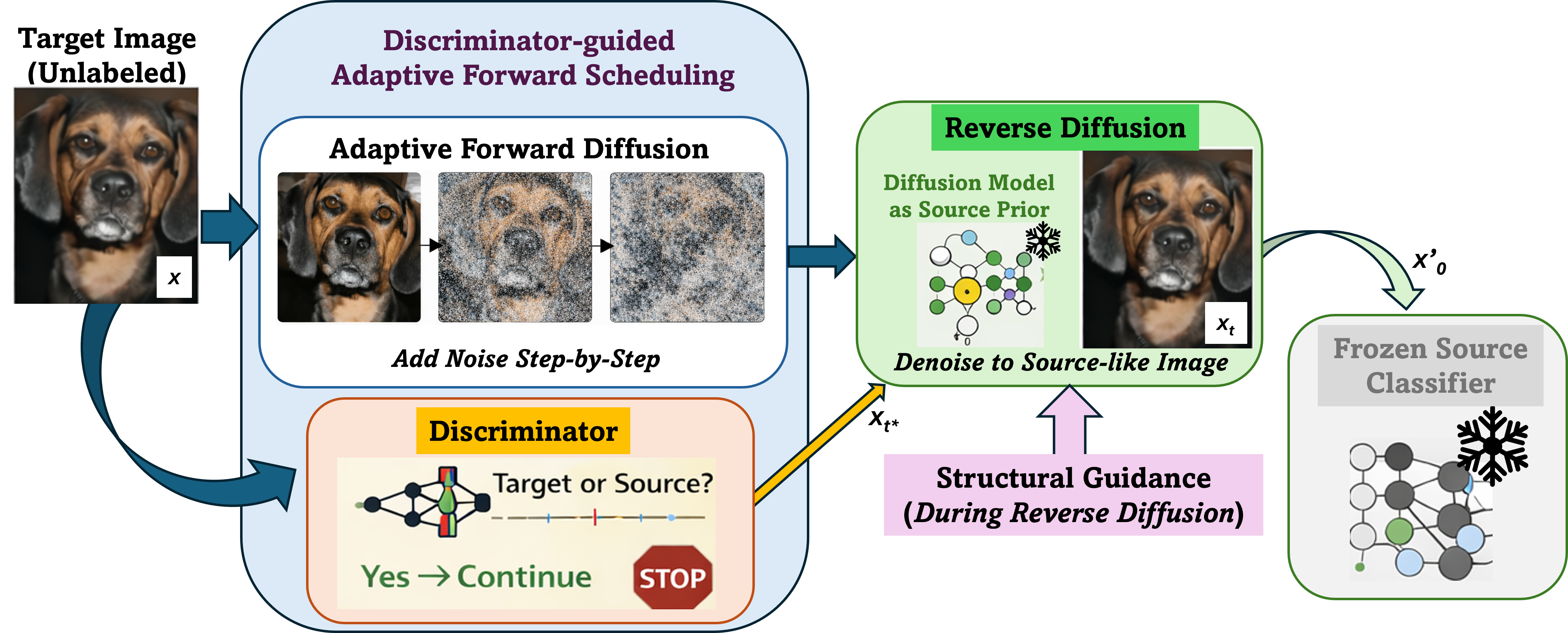}
    \caption{ Overview of the proposed discriminator-guided adaptive diffusion framework. The method operates entirely at test time by combining a pretrained diffusion model with a discriminator. Forward diffusion incrementally injects noise into a target input, and the discriminator monitors the evolution of domain-specific cues to determine an adaptive stopping point $t^*$. From the selected latent state $x_{t^*}$, reverse diffusion generates a source-aligned image, guided to preserve the input’s low-frequency structure. The adapted image is finally classified using a frozen source-trained classifier.}
    \label{fig:overview}
    \vspace{-1.8em}
\end{figure}

\section{Proposed Method}
\label{sec:method}

Our method adapts target domain images by leveraging a pretrained diffusion model together with a discriminator-guided adaptive forward noising mechanism. Given an unlabeled target image, we progressively perturb the input via forward diffusion while a discriminator evaluates each intermediate state to determine whether target-domain cues remain. The forward process is halted adaptively at a timestep $t^*$ once the discriminator can no longer reliably identify the sample as belonging to the target domain. Starting from the selected state $x_{t^*}$, reverse diffusion reconstructs a source-aligned image. During this reverse process, we incorporate a structural guidance mechanism that preserves the low-frequency structure of the input image. The resulting adapted image $x'_0$ is then classified by a frozen source-trained classifier. An overview of the method is shown in Fig.~\ref{fig:overview}.

Importantly, the proposed framework does not perform any training or parameter optimization at test time.
The diffusion model and classifier are pretrained on the source domain and remain fixed during inference, while the discriminator is trained offline prior to deployment and kept frozen at test time. The entire procedure consists of a single inference-time pipeline applied independently to each test image, where adaptivity arises solely from the sample-specific stopping of the forward diffusion process.
\vspace{-1em}

\subsection{Frozen Diffusion Model as Source Prior}
\label{subsec:DDPM}

A Denoising Diffusion Probabilistic Model (DDPM)~\cite{ho2020denoising} is a widely used class of generative models that learns a data distribution through a forward noising process and a learned reverse denoising process. In the forward process, Gaussian noise is gradually added to an image $\mathbf{x}_0$ over $T$ timesteps, producing a sequence of increasingly noisy latents $\mathbf{x}_1, \dots, \mathbf{x}_T$. The reverse process is parameterized by a neural network trained to invert this corruption, progressively denoising $\mathbf{x}_T$ towards samples drawn from the training distribution.

In our framework, we employ a DDPM pretrained on source-domain data. Because the reverse process has learned to map noisy inputs toward the source distribution, the diffusion model serves as an effective source prior. When initialized from a suitably noised target image, the reverse diffusion process reconstructs a source-aligned sample consistent with the learned source statistics. Importantly, the diffusion model remains fixed during adaptation and is not updated using target-domain data. While this source prior is powerful, applying it with an inappropriate amount of forward noising can lead to unnecessary semantic degradation. We address this challenge by adaptively selecting the stopping point of the forward diffusion process using a discriminator-guided scheduling strategy, described below.
\vspace{-1em}

\subsection{Discriminator-guided Adaptive Forward Scheduling}
\label{subsec:adaptation}

Although the diffusion noise schedule itself is fixed, the stopping point of the forward diffusion process is selected adaptively on a per-sample basis. Specifically, a discriminator determines when sufficient perturbation has been applied to suppress target-domain cues, thereby selecting the stopping timestep $t^*$. Our core hypothesis is that a corrupted target domain image still contains the essential semantic information of the original image, but is distorted by domain-specific artifacts. By treating these artifacts as noise, we aim to inject only the minimal amount of perturbation required to neutralize them, avoiding unnecessary degradation of semantic content.

\begin{figure}[t!]
    \centering
    \includegraphics[width=0.87\linewidth]{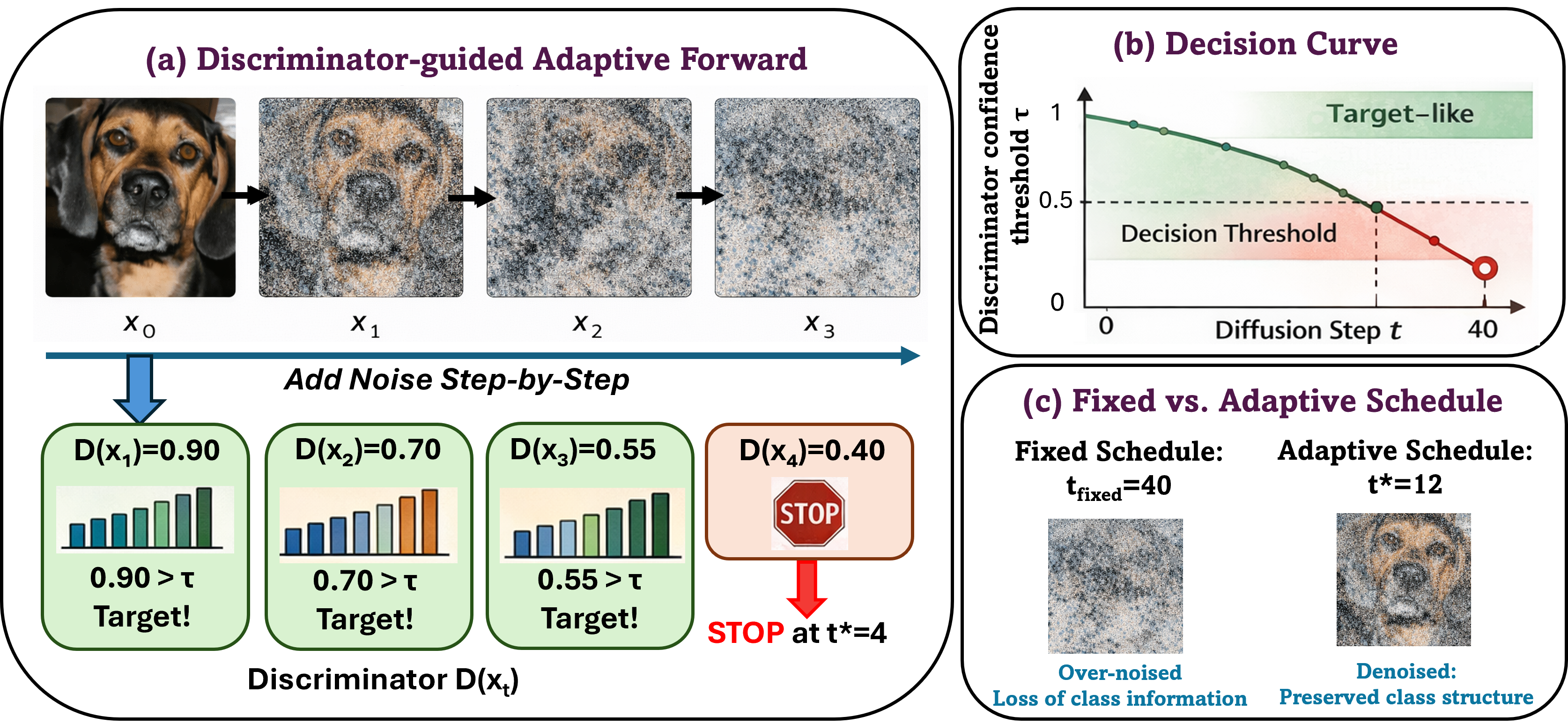}
    \caption{(a) Illustration of discriminator-guided adaptive forward scheduling. Forward diffusion progressively perturbs an unlabeled target image $x_0$, while a domain discriminator evaluates $D(x_t)$ at each step. Diffusion continues while $D(x_t)$ $\ge$ $\tau$ and is halted at the adaptive stopping timestep $t^*$ once $D(x_{t^*})$ $<$ $\tau$.
(b) Conceptual decision curve of \textbf{discriminator confidence} versus diffusion step, with the stopping point determined by the threshold $\tau$.
(c) Comparison of fixed and adaptive schedules, showing that fixed schedules may over-noise inputs and remove class-discriminative information, whereas adaptive scheduling halts earlier to better preserve class-discriminative structure.}
    \label{fig2}
    \vspace{-1.8em}
\end{figure}

Let $\mathbf{x}_{\text{target}}$ denote a target image, which we identify with $\mathbf{x}_0$ in the diffusion notation. During adaptation, we iteratively apply the forward diffusion process to obtain intermediate states $\mathbf{x}_t$. At each timestep $t$, a domain discriminator $D(\mathbf{x}_t)$ predicts whether the noised image still exhibits target-domain characteristics. Forward diffusion continues while the discriminator confidently identifies target-domain cues and is halted at timestep $t^*$ once this \textit{confidence} drops below a predefined threshold $\tau$. This procedure adaptively determines the amount of perturbation applied to each input. Fig.~\ref{fig2} illustrates this scheduling mechanism.
\vspace{-2em}

\subsubsection{Training the Discriminator.}
The discriminator is trained to distinguish between noisy images originating from the target domain and noisy images drawn from the source distribution. Concretely, it solves a binary classification task on two types of inputs:
%\begin{itemize}
    %\item 
    \textit{(1) Noisy target samples}: target images $\mathbf{x}_{\text{target}}$ corrupted with Gaussian noise at a randomly sampled forward timestep $t$, producing inputs $\mathbf{x}_t$.
    %\item 
    \textit{(2) Noisy source-like samples}: clean images sampled from the pretrained diffusion model (approximating the source distribution) and similarly corrupted with Gaussian noise at a random timestep $t$.
%\end{itemize}
By training on inputs with varying noise levels, the discriminator learns to identify domain-specific cues independently of noise intensity, effectively estimating how much perturbation is required for a target image to become indistinguishable from a source-like sample.
\vspace{-1em}

\subsubsection{Driving the Forward Process.}
During test-time adaptation, we apply a forward diffusion step-by-step and query the discriminator at each timestep. The forward process is halted at timestep $t^*$ once the discriminator no longer confidently identifies the image as target-domain. This adaptive stopping criterion controls the amount of noise injected into each sample and serves as the sole output of the scheduling stage.

\vspace{-1em}
\subsection{Reverse Denoising}
\label{subsec:reverse}
Once the adaptive stopping timestep $t^*$ has been determined, we initialize the reverse diffusion process from the selected intermediate state $\mathbf{x}_{t^*}$. Because the diffusion model is pretrained on the source domain, and kept frozen during adaptation, the reverse process progressively denoises $\mathbf{x}_{t^*}$ toward a sample that is consistent with the source data distribution. This reverse denoising step reconstructs an adapted image $\mathbf{x}'_0$ by filling in missing details using source-consistent structure.

\vspace{-1em}
%\subsection{Guidance-free conditioning}
\subsection{Structural Guidance}
\label{subsec:guidance}

While adaptive forward scheduling controls how much noise is injected, reverse denoising must also preserve the global structure of the input image. Relying solely on the diffusion model’s learned prior can introduce deviations from the original signal, particularly under severe corruptions. To encourage structural fidelity, we incorporate an image-based structural guidance mechanism during the reverse diffusion process. This guidance is applied only at test-time and does not require training a conditional diffusion model. The conditioning signal is obtained directly from the corrupted input image via a low-pass filtering operation, $c = LPF(x_c)$.
At each reverse diffusion step, we compute an estimate of the clean image $\hat{x}_0$ and define a structural reconstruction loss that penalizes discrepancies between the low-frequency components of $\hat{x}_0$ and the conditioning signal: $\mathcal{E}_{\text{struct}} = \frac{1}{2}\left\| LPF(\hat{x}_0) - c \right\|_2^2$.
The gradient of this loss with respect to the current diffusion state $x_t$ is incorporated into the reverse diffusion update. By constraining only low-frequency content, this guidance preserves global image structure while allowing the diffusion model to synthesize high-frequency details, improving robustness for classification.

\vspace{-1em}
\subsection{Frozen Source Classifier}
\label{subsec:classifier}
After adaptation, the reconstructed image $\mathbf{x}'_0$ is passed to a source-trained classifier to obtain the final prediction. The classifier is pretrained on the source domain and remains completely frozen during test-time adaptation. No classifier parameters are updated, and no additional test-time optimization or ensembling is performed. This design ensures that all adaptation is achieved through input transformation, making the method strictly source-free and classifier-free.
\vspace{-1em}

\section{Experimental Analysis}
\label{sec:experiments}

\subsubsection{Datasets and Evaluation Metric.}
Following prior work~\cite{gao2022back,nie2022diffusion}, we evaluate our method on ImageNet-C (IN-C)~\cite{hendrycks2019benchmarking,Schneider2020CovShift}, a standard benchmark for robustness in large-scale image classification. ImageNet-C consists of 50{,}000 validation images from ImageNet subjected to synthetic yet naturalistic corruptions, spanning four broad categories, noise, blur, digital artifacts, and weather effects, for a total of 15 corruption types that effectively function as separate target domains in practice. 
In line with earlier works, we assess classification robustness using top-1 accuracy at the highest corruption severity (severity level 5), and all methods are evaluated under identical experimental settings to ensure fair comparison. In addition, when analyzing the behavior of the domain discriminator in our ablation studies, we report the F1-score to evaluate its binary discrimination performance.
\vspace{-1em}

\subsubsection{Implementation Details.}
Our method builds on a DDPM pretrained on the ImageNet source domain. At inference time, we evaluate the diffusion model using a rescaled schedule with $T = 100$ timesteps and a linear noise schedule. The diffusion model remains frozen throughout adaptation.
The domain discriminator $D(\cdot)$ is trained to distinguish between noisy target-domain samples and noisy source-like samples generated by the pretrained diffusion model. It takes noisy images as input and outputs a binary prediction indicating whether a sample is target-like. The discriminator is trained using binary cross-entropy loss with the Adam optimizer, a learning rate of $2 \times 10^{-5}$, and for 10 epochs.
Adaptive scheduling uses a fixed discriminator \textit{confidence threshold} $\tau = 0.5$ (notice that although $\tau$ is fixed, the sample-dependent stopping time varies across images and corruption types and $\tau = 0.5$ simply corresponds to the discriminator reaching chance-level confidence). $\tau$ is fixed globally and not tuned per corruption or target domain, preserving the source-free test-time setting.
Structural guidance is incorporated during reverse diffusion as a gradient-based correction weighted by a guidance scale $\lambda$. We set $\lambda = 6$ for all experiments, in line with~\cite{gao2022back}.
For classification, we use a Swin-Tiny Transformer~\cite{swin2021transformer} pretrained on ImageNet and keep it frozen during adaptation to ensure fair comparison with prior works~\cite{gao2022back,ho2020denoising}. All experiments are conducted on a single NVIDIA RTX 4090 GPU with a batch size of 8.

%%%%%%%%%%%%%%%%%%
\begin{table*}[t]
\centering
\scriptsize
\setlength{\tabcolsep}{4pt}
\renewcommand{\arraystretch}{1.1}
\caption{Comparison of source-free test-time adaptation methods in terms of Top-1 Accuracy (ACC) and other characteristics. Tent~\cite{wang2021tent}, MEMO~\cite{zhang2021memo}, DiffPure~\cite{nie2022diffusion} results are taken from DDA~\cite{gao2022back}. Best shown in \textbf{black}.}
\resizebox{0.88\textwidth}{!}{%
\begin{tabular}{lccccc}
\toprule
\textbf{Method} 
& \textbf{ACC} 
& \textbf{Adaptation} 
& \textbf{Diffusion?} 
& \textbf{Per-sample?} 
& \textbf{Frozen Classifier?} \\
\midrule

Source-only
& 0.33 
& None 
& \ding{55} 
& \ding{55} 
& \ding{51} \\

\midrule

Tent~\cite{wang2021tent} 
& 0.30 
& Model 
& \ding{55} 
& \ding{55} 
& \ding{55} \\

MEMO~\cite{zhang2021memo} 
& 0.29
& Model 
& \ding{55} 
& \ding{51} 
& \ding{55} \\

\midrule

DiffPure~\cite{nie2022diffusion} 
& 0.25 
& Input 
& \ding{51} 
& \ding{55} 
& \ding{51} \\

DDA~\cite{gao2022back} 
& 0.38 
& Input 
& \ding{51} 
& \ding{55} 
& \ding{51} \\

\rowcolor{blue!10}
Ours 
& 0.36
& Input 
& \ding{51} 
& \ding{51} 
& \ding{51} \\

\rowcolor{blue!10}
Ours + Guidance 
& \textbf{0.39}
& Input 
& \ding{51} 
& \ding{51} 
& \ding{51} \\

\bottomrule
\end{tabular}}

\label{method_comparison}
\vspace{-2em}
\end{table*}

%%%%%%%%%%%%%%%%%%

%%%%%% Detailed comparisons %%%%%%%

\begin{table}[hb]
\centering
\scriptsize
\setlength{\tabcolsep}{3pt}
\caption{Performance comparisons grouped into four corruption families. Each group reports top-1 accuracy along with the average over corruptions in the group. Abbreviations: G=Guidance, Gaus=Gaussian Noise, Shot=Shot Noise, Imp=Impulse Noise, Def=Defocus Blur, Mot=Motion Blur, Bri=Brightness, Cont=Contrast, Ela=Elastic Transform, Pix=Pixelate.}
\renewcommand{\arraystretch}{1.0}

\begin{subtable}{0.48\linewidth}
\centering
\begin{tabular}{lcccc|c}
\toprule
\textbf{Blur} & Def & Glass & Mot & Zoom & \textbf{Avg} \\
\midrule
Source & 0.21 & 0.10 & 0.24 & 0.26 & 0.20 \\
DiffPure & 0.12 & \textbf{0.18} & 0.15 & 0.19 & 0.16 \\
DDA & 0.21 & 0.16 & 0.23 & 0.25 & 0.21 \\
\rowcolor{blue!15} Ours & \textbf{0.22} & 0.17 & 0.24 & 0.25 & 0.22 \\
\rowcolor{blue!15} Ours+G & \textbf{0.22} & 0.17 & \textbf{0.25} & \textbf{0.27} & \textbf{0.23} \\
\bottomrule
\end{tabular}
\end{subtable}
\hfill
\begin{subtable}{0.48\linewidth}
\centering
\begin{tabular}{lccc|c}
\toprule
\textbf{Weather} & Snow & Frost & Fog & \textbf{Avg} \\
\midrule
Source & 0.36 & 0.42 & 0.41 & 0.40 \\
DiffPure & 0.19 & 0.32 & 0.08 & 0.20 \\
DDA & 0.35 & \textbf{0.46} & 0.38 & 0.40 \\
\rowcolor{blue!15} Ours & \textbf{0.36} & \textbf{0.46} & \textbf{0.43} & \textbf{0.42} \\
\rowcolor{blue!15} Ours+G & \textbf{0.36} & 0.44 & \textbf{0.43} & \textbf{0.41} \\
\bottomrule
\end{tabular}
\end{subtable}

\vspace{0.3cm}

\begin{subtable}{0.48\linewidth}
\centering
\begin{tabular}{lccccc|c}
\toprule
\textbf{Digital} & Bri & Cont & Ela & Pix & JPEG & \textbf{Avg} \\
\midrule
Source & 0.66 & 0.34 & 0.22 & 0.32 & 0.51 & 0.41 \\
DiffPure & 0.57 & 0.01 & 0.33 & 0.34 & 0.49 & 0.35 \\
DDA & 0.63 & 0.29 & 0.34 & 0.50 & 0.51 & 0.45 \\
\rowcolor{blue!15} Ours & 0.65 & \textbf{0.35} & \textbf{0.37} & \textbf{0.57} & \textbf{0.55} & \textbf{0.50} \\
\rowcolor{blue!15} Ours+G & \textbf{0.67} & \textbf{0.35} & \textbf{0.37} & 0.53 & 0.52 & 0.49 \\
\bottomrule
\end{tabular}
\end{subtable}
\hfill
\begin{subtable}{0.48\linewidth}
\centering
\begin{tabular}{lccc|c}
\toprule
\textbf{Noise} & Gaus & Shot & Imp & \textbf{Avg} \\
\midrule
Source & 0.30 & 0.29 & 0.28 & 0.29 \\
DiffPure & 0.24 & 0.23 & 0.23 & 0.23 \\
DDA & \textbf{0.50} & \textbf{0.50} & \textbf{0.51} & \textbf{0.50} \\
\rowcolor{blue!15} Ours & 0.24 & 0.24 & 0.23 & 0.24 \\
\rowcolor{blue!15} Ours+G & 0.39 & 0.40 & 0.40 & 0.40 \\
\bottomrule
\end{tabular}
\end{subtable}

\label{tab:corruption_groups}
\vspace{-2em}
\end{table}

\vspace{-1em}
\subsection{Results}
\subsubsection{Comparisons with the State-of-the-art.}
Tab.~\ref{method_comparison} compares SOTA test-time SFUDA methods in terms of both overall Top-1 accuracy and key methodological characteristics. Our method with guidance achieves the highest overall accuracy, slightly outperforming DDA~\cite{gao2022back} while uniquely enabling per-sample adaptive scheduling. This comparison highlights the practical advantages of our approach beyond aggregate performance.

We further provide a detailed comparison between the source-only baseline and SOTA input-adaptation diffusion-based methods (i.e., DiffPure~\cite{nie2022diffusion} and DDA~\cite{gao2022back}) against our method and its guided variant across individual corruption families (Tab.~\ref{tab:corruption_groups}).
Overall, our method exhibits stable and competitive performance, avoiding the severe degradation observed in DiffPure~\cite{nie2022diffusion} while remaining competitive with DDA~\cite{gao2022back}. DiffPure~\cite{nie2022diffusion}, which is primarily designed for adversarial perturbations, consistently underperforms the source-only baseline on natural corruptions, confirming that a fixed and shallow diffusion depth is insufficient for handling diverse corruption types. In contrast, our method maintains or improves accuracy over the source-only model across blur, weather, and digital corruptions.

On \textbf{blur corruptions}, our method achieves the strongest average performance. In particular, the guided variant yields the highest average accuracy, with improvements on motion blur and zoom blur. Both variants of our method outperform DDA~\cite{gao2022back} on defocus blur and motion blur, indicating improved robustness under spatially distributed degradations. This suggests that adaptive stopping helps preserve class-relevant structure while avoiding excessive smoothing.
On \textbf{weather corruptions}, our method without guidance achieves the highest average accuracy, outperforming both the source-only baseline and DDA~\cite{gao2022back}. Improvements are especially pronounced on fog and snow, where adaptive scheduling enables effective removal of corruption without degrading semantic content. The guided variant remains competitive, indicating that structural guidance does not harm performance in this regime.
\textbf{Digital corruptions} further highlight the benefit of adaptive scheduling. Our method without guidance achieves the highest average accuracy in this group, with consistent gains on contrast, pixelation, and JPEG compression. These results suggest improved semantic preservation, as overly aggressive denoising tends to distort global image statistics that are critical for these corruptions. The guided variant remains competitive but slightly underperforms the unguided version on average, reflecting a trade-off between structural constraint and flexibility in reconstruction.
On \textbf{noise-based corruptions}, DDA~\cite{gao2022back} achieves the strongest performance, reflecting its aggressive reverse denoising strategy. Our method without guidance performs conservatively in this regime, as the discriminator often halts the forward process early when noise is already present in the input. When augmented with structural guidance, our approach substantially improves performance (from 0.24 to 0.40 on average), though it remains below DDA~\cite{gao2022back}. This behavior reflects a deliberate trade-off: \textit{our method prioritizes avoiding over-denoising, which can be detrimental outside pure noise settings.}

Beyond accuracy, our method offers practical advantages in adaptivity and deployment. Unlike DDA~\cite{gao2022back}, which employs a fixed diffusion range together with additional guidance and ensembling mechanisms, our approach automatically determines the effective diffusion depth on a per-sample basis using a discriminator, eliminating the need for manual diffusion-depth selection. While diffusion-based SFUDA methods keep models frozen at test time, our per-sample adaptive scheduling enables robust input-level adaptation without introducing additional optimization or ensembling procedures. As a result, whereas DDA~\cite{gao2022back} excels on pure noise corruptions, our method achieves more balanced robustness across blur, weather, and digital corruptions. 
%By avoiding over-denoising and requiring minimal tuning, our approach provides a stable and practical solution for test-time adaptation under diverse natural corruptions.
By avoiding over-denoising and reducing the need for corruption-specific tuning, particularly manual diffusion-depth selection, our approach provides a stable and practical solution for TTA under diverse natural corruptions.

\begin{figure}[t!]
    \centering
    \includegraphics[width=0.75\linewidth]{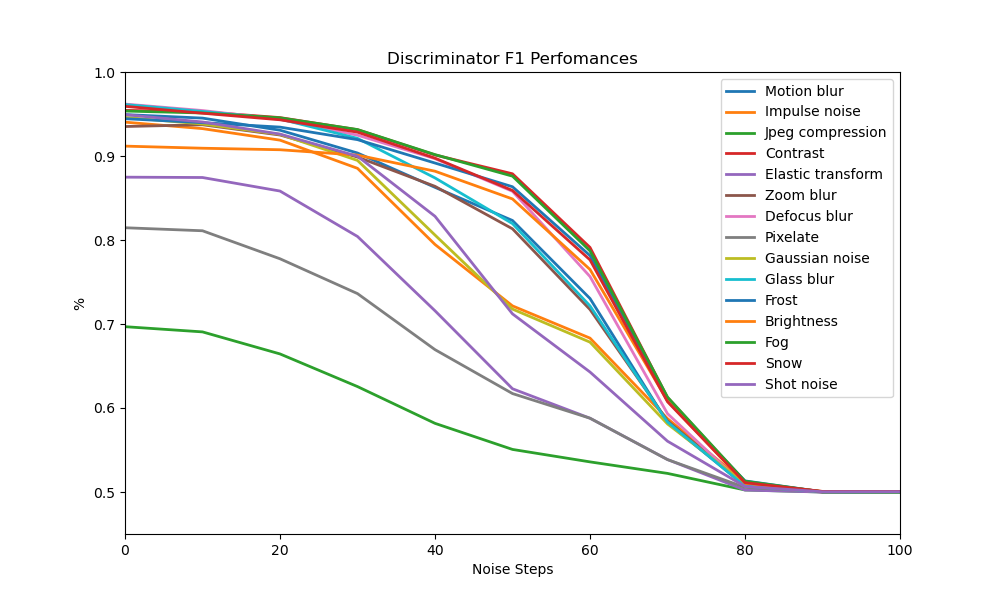}
    \caption{Discriminator performance as a function of forward diffusion depth for different corruption types. The discriminator measures its ability to distinguish noised target samples from source-like samples at each timestep. Diffusion steps are rescaled from 1000 to 100 for visualization. Different decay rates across corruptions indicate that domain-specific cues are removed at different speeds, motivating adaptive per-sample stopping of the forward diffusion process.}
    \vspace{-2em}
    \label{all_disc_perf}
\end{figure}

\vspace{-1em}
\subsubsection{Discriminator Behavior Across Corruptions.}
Fig. \ref{all_disc_perf} analyzes the behavior of the domain discriminator across different corruption types as a function of the forward diffusion depth. We report the discriminator performance when distinguishing noised target samples from source-like samples at each forward timestep. Across all corruptions, discriminator performance degrades monotonically as noise increases, confirming that forward diffusion progressively removes domain-specific cues. However, the rate of degradation varies significantly across corruption families. Noise-based corruptions (Gaussian, Shot, Impulse) exhibit a slower decline in discriminability, indicating that substantial diffusion is required before target-specific artifacts become indistinguishable from source-like noise. In contrast, blur, weather, and digital corruptions lose discriminative cues much earlier, often within the first half of the diffusion range.

This behavior directly motivates our adaptive scheduling strategy. Rather than applying a fixed diffusion depth, our method halts the forward process once the discriminator can no longer reliably identify the target domain. As a result, inputs affected by structural or semantic corruptions (e.g., blur, fog, JPEG compression) are adapted with fewer diffusion steps, preserving class-relevant content, while inputs dominated by noise naturally undergo deeper diffusion. Importantly, the convergence of discriminator performance across corruptions at large timesteps reflects the expected collapse toward noise, reinforcing the need to avoid excessive diffusion. These observations explain the balanced performance of the proposed method across corruption families and its robustness to over-denoising compared to fixed-schedule approaches.

%%%%%%%%%%%%%%%%%Adaptive Stopping Behavior%%%%%%%%%%%%%%%%%%%%

\begin{figure}[t!]
    \centering

    % -------- Noise --------
    \begin{subfigure}[t]{0.45\linewidth}
        \centering
        \includegraphics[width=\linewidth]{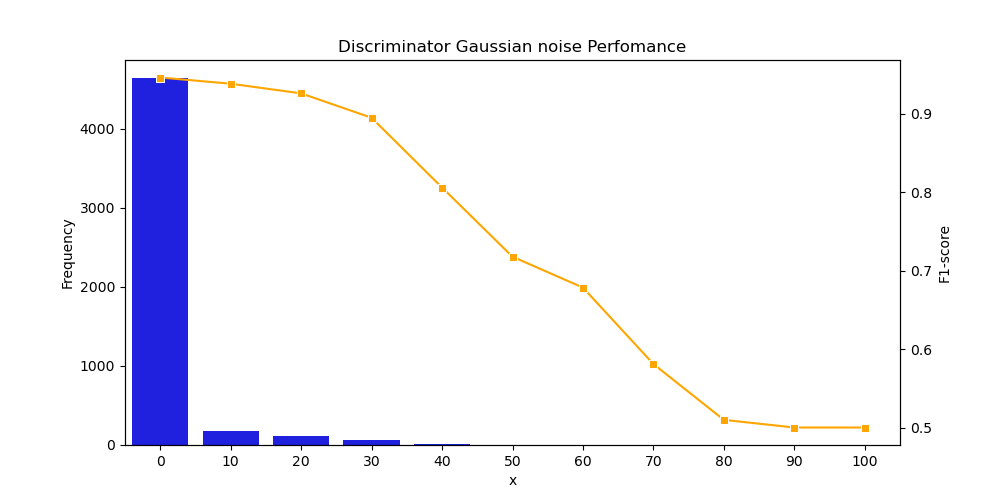}
        %\caption{Gaussian noise}
    \end{subfigure}
    \hfill
    % -------- Blur --------
    \begin{subfigure}[t]{0.45\linewidth}
        \centering
        \includegraphics[width=\linewidth]{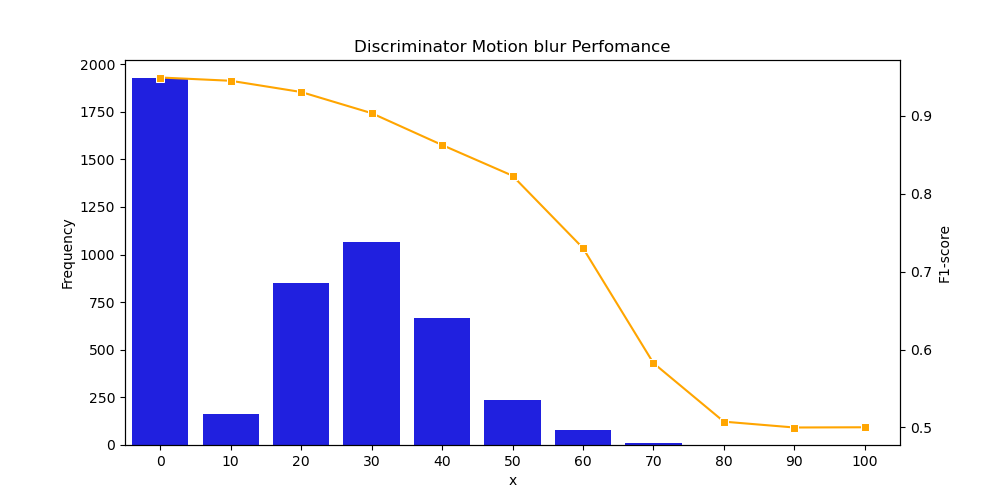}
        %\caption{Motion blur}
    \end{subfigure}

    %\vspace{-0.2em}

    % -------- Weather --------
    \begin{subfigure}[t]{0.45\linewidth}
        \centering
        \includegraphics[width=\linewidth]{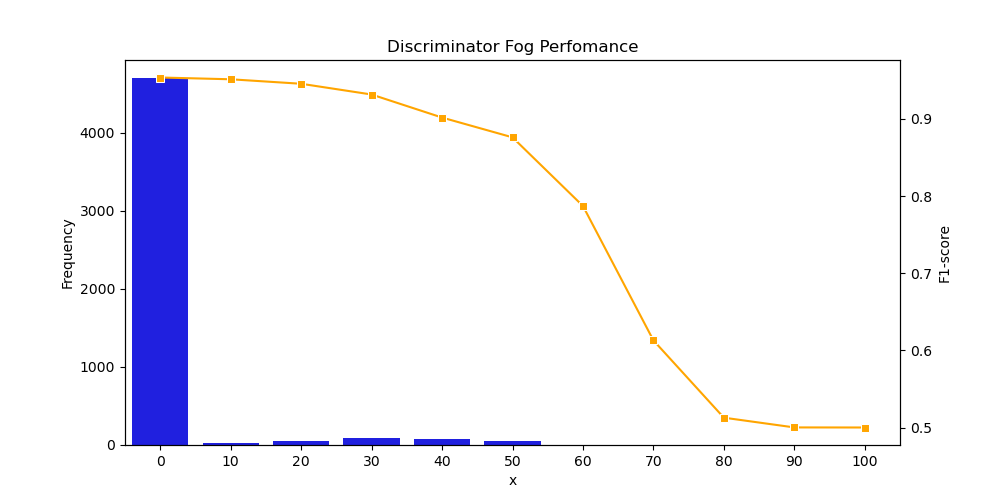}
        %\caption{Fog}
    \end{subfigure}
    \hfill
    % -------- Digital --------
    \begin{subfigure}[t]{0.45\linewidth}
        \centering
        \includegraphics[width=\linewidth]{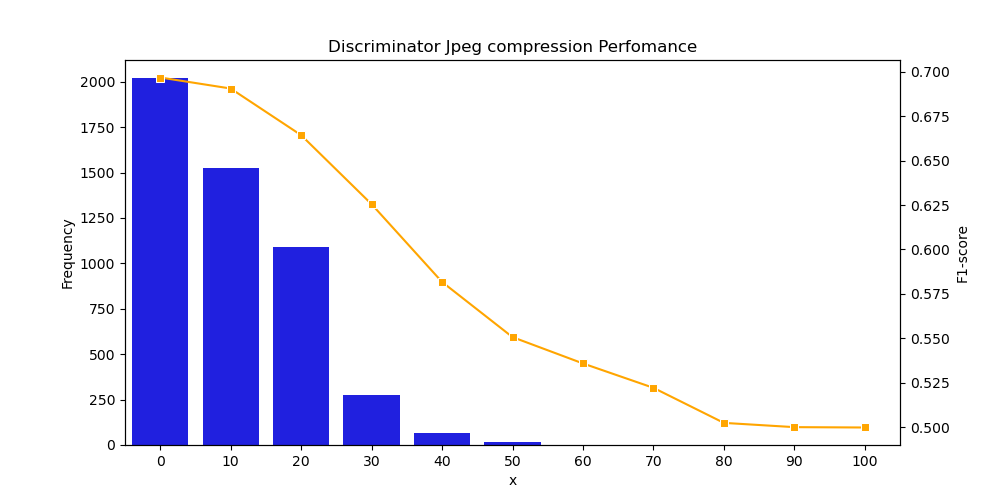}
        %\caption{JPEG compression}
    \end{subfigure}

    \caption{Representative examples of discriminator-guided adaptive stopping behavior, with one corruption selected from each corruption family (noise, blur, weather, and digital). Each subfigure shows the distribution of adaptive stopping timesteps $t^*$ (blue histogram) together with the discriminator performance across forward diffusion steps (orange curve). Results for all 15 corruption types are reported in the Supp. Mat.}
    \vspace{-1.8em}
    \label{fig:ind_perf}
\end{figure}

\vspace{-1em}
\subsubsection{Adaptive Stopping Behavior of the Discriminator.}
While Fig.~\ref{all_disc_perf} characterizes the global decay of discriminator performance as diffusion increases across all corruption types, Fig.~\ref{fig:ind_perf} shows how this signal is operationalized into per-sample adaptive stopping decisions. We visualize one representative corruption from each ImageNet-C family (Gaussian noise, motion blur, fog, and JPEG compression), with results for all 15 corruption types reported in the Supp. Mat. Each subplot reports the distribution of adaptive stopping timesteps $t^*$ together with the corresponding discriminator performance as a function of forward diffusion depth.

Gaussian noise represents noise-based corruptions, where domain-specific artifacts persist under moderate diffusion and require deeper forward perturbation before the discriminator becomes uncertain. Motion blur exemplifies spatially structured blur corruptions, for which domain cues are removed more rapidly, and adaptive stopping occurs at earlier diffusion steps. Fog represents challenging weather corruptions with globally distributed effects, where aggressive diffusion can easily distort semantic content if not carefully controlled. JPEG compression is a representative digital corruption, where adaptive stopping typically occurs at intermediate diffusion depths to preserve global image statistics while suppressing compression artifacts.

Across these representative cases, stopping distributions concentrate near the diffusion depth at which the discriminator performance degrades toward chance level, indicating that the adaptive stopping criterion is driven by the discriminator’s ability to distinguish target-domain structure rather than by a fixed or heuristic diffusion schedule. The variation in the location and spread of the stopping distributions across corruption families reflects the differing amounts of diffusion required to neutralize domain-specific cues, and helps explain the balanced robustness observed across corruption types in Tab.~\ref{tab:corruption_groups}. Overall, these representative examples demonstrate that the proposed discriminator-guided adaptive scheduling selects the diffusion depth in a corruption-aware manner, enabling effective removal of domain-specific artifacts while avoiding unnecessary over-diffusion across diverse corruption families.

%%%%%%%%%%%%%%%%%%%%%%%Confidence Threshold Ablation%%%%%%%%%%%%%%%%%%%%%%%%%

\begin{table}[t!]
\centering
\scriptsize
\setlength{\tabcolsep}{3pt}
\renewcommand{\arraystretch}{1.0}
\caption{Ablations of the discriminator confidence threshold $\tau$ without guidance. Each group reports the top-1 accuracy along with the average over corruptions in the group.}
\begin{subtable}{0.48\linewidth}
\centering
\begin{tabular}{lcccc|c}
\toprule
\textbf{$\tau$} & Def & Glass & Mot & Zoom & \textbf{Avg} \\
\midrule
0.10 & \textbf{0.22} & 0.10 & \textbf{0.24} & 0.24 & 0.20 \\
0.30 & \textbf{0.22} & 0.10 & \textbf{0.24} & 0.24 & 0.20 \\
\rowcolor{blue!15} 0.50 & \textbf{0.22} & \textbf{0.17} & \textbf{0.24} & \textbf{0.25} & \textbf{0.22} \\
0.70 & \textbf{0.22} & 0.10 & \textbf{0.24} & 0.23 & 0.20 \\
\bottomrule
\end{tabular}
\end{subtable}
\hfill
\begin{subtable}{0.48\linewidth}
\centering
\begin{tabular}{lccc|c}
\toprule
\textbf{$\tau$} & Snow & Frost & Fog & \textbf{Avg} \\
\midrule
0.10 & 0.34 & 0.45 & \textbf{0.43}  & 0.41 \\
0.30 & 0.34 & 0.45 & \textbf{0.43}  & 0.41 \\
\rowcolor{blue!15} 0.50 & \textbf{0.35} & \textbf{0.46} & \textbf{0.43}  & \textbf{0.42} \\
0.70 & 0.30 &  0.43 & \textbf{0.43} & 0.39 \\
\bottomrule
\end{tabular}
\end{subtable}

\vspace{0.2cm}

\begin{subtable}{0.48\linewidth}
\centering
\begin{tabular}{lccccc|c}
\toprule
\textbf{$\tau$} & Bri & Cont & Ela & Pix & JPEG & \textbf{Avg} \\
\midrule
0.10 & 0.65 & 0.35 & 0.24 & 0.27 & 0.48 & 0.40 \\
0.30 & 0.65 & 0.35 & 0.29 & 0.29 & 0.52 & 0.42 \\
\rowcolor{blue!15} 0.50 & 0.65 & 0.35 & \textbf{0.37} & \textbf{0.57} & \textbf{0.55} & \textbf{0.50} \\
0.70 & 0.64 & \textbf{0.36} & 0.24 & 0.40 & 0.50 & 0.43 \\
\bottomrule
\end{tabular}
\end{subtable}
\hfill
\begin{subtable}{0.48\linewidth}
\centering
\begin{tabular}{lccc|c}
\toprule
\textbf{$\tau$} & Gaus & Shot & Imp & \textbf{Avg} \\
\midrule

0.10 & \textbf{0.24} & \textbf{0.24} & 0.22  & 0.23 \\
0.30 & \textbf{0.24} & \textbf{0.24} & \textbf{0.23}  & \textbf{0.24} \\
\rowcolor{blue!15} 0.50 & \textbf{0.24} & \textbf{0.24} & \textbf{0.23} & \textbf{0.24} \\
0.70 & 0.22 & 0.22 & 0.20 & 0.21 \\

\bottomrule
\end{tabular}
\end{subtable}

\vspace{-2em}
\label{tab:disc_ablations}
\end{table}

\vspace{-1em}
\subsubsection{Ablations on the discriminator confidence threshold $\tau$.}
We analyze the sensitivity of our method to the discriminator confidence threshold $\tau$, which controls the stopping criterion for the forward diffusion process (Tab.~\ref{tab:disc_ablations}).
Overall, the method exhibits stable performance across a broad range of confidence thresholds. In particular, $\tau = 0.50$ consistently achieves the highest or near-highest average accuracy across blur, weather, and digital corruptions, while maintaining competitive performance on noise corruptions. Lower confidence thresholds tend to halt diffusion prematurely, leading to insufficient removal of domain-specific artifacts, whereas higher confidence thresholds introduce excessive perturbation that degrades semantic content. These results indicate that the proposed adaptive scheduling is not overly sensitive to the choice of $\tau$, and that a mid-range confidence threshold (i.e., $\tau$ = 0.50) provides a favorable balance between domain shift removal and semantic preservation.

%%%%%%%%%%%%%%t-SNE%%%%%%%%%%%%%%%%%%%%
\vspace{-1em}
\subsubsection{Feature Visualization.}
Additional t-SNE visualizations of feature representations across corruption families are provided in the Supp. Mat. These visualizations suggest that noise-based corruptions tend to form clusters that are more separated from the source domain, whereas blur, weather, and digital corruptions exhibit closer alignment. While not used for quantitative analysis or methodological decisions, these plots provide intuition for the corruption-dependent behavior observed in adaptive diffusion.

%%%%%%%%%%%%%%t-SNE%%%%%%%%%%%%%%%%%%%%

\vspace{-1em}
\section{Discussions}
\label{sec:discussions}
Results highlight a clear distinction between the behavior of DDA~\cite{gao2022back} and our diffusion scheduling. DDA achieves its strongest performance on noise-based corruptions, where an aggressive fixed diffusion range effectively removes additive noise, but this strategy can be suboptimal for non-noise corruptions such as blur, weather, and digital artifacts. In contrast, our method applies the minimum perturbation required to suppress domain-specific cues by adaptively selecting the stopping timestep on a per-sample basis, thereby avoiding excessive diffusion that would remove class-discriminative information. This design yields more balanced performance across blur, weather, and digital corruptions, where over-denoising is particularly harmful. Adaptive scheduling is especially relevant in realistic source-free test-time adaptation scenarios with substantial inter- and intra-class variability, where a single fixed diffusion schedule cannot be optimal across all inputs. Finally, structural guidance and adaptive scheduling play complementary roles: the former stabilizes reverse diffusion and preserves global image structure, while the latter controls the strength of forward perturbation, jointly improving robustness without additional tuning complexity.

\vspace{-1em}
\section{Conclusion}
\label{sec:conc}

We introduced a discriminator-guided adaptive diffusion scheduling framework for source-free test-time adaptation under image corruptions. Unlike prior diffusion methods with fixed schedules, our approach selects the diffusion depth per sample, enabling corruption-aware adaptation without modifying source-trained models. This balances domain shift removal with preservation of class-discriminative features, yielding robust performance across diverse corruptions. The method is fully source-free, operates on individual samples, and requires no corruption-specific tuning or retraining, making it suitable for heterogeneous and unknown corruption settings. While fixed schedules remain effective for noise-based corruptions, adaptive scheduling better handles non-noise types such as blur, weather, and digital artifacts. Future work will explore hybrid strategies combining adaptive scheduling with corruption-aware guidance.

\subsubsection{\ackname} We acknowledge the financial support  of the PNRR project FAIR - Future AI Research (PE00000013),  
under the NRRP MUR program funded by the NextGenerationEU.

\section*{SUPPLEMENTARY MATERIAL}

\begin{figure*}[h!]
\centering
\setlength{\tabcolsep}{2pt}

% -------- Row 1: Noise --------
\begin{subfigure}[t]{0.3\textwidth}
    \centering
    \includegraphics[width=\linewidth]{images/eval_gaussian_noise.png}
    \caption{Gaussian}
\end{subfigure}
\begin{subfigure}[t]{0.3\textwidth}
    \centering
    \includegraphics[width=\linewidth]{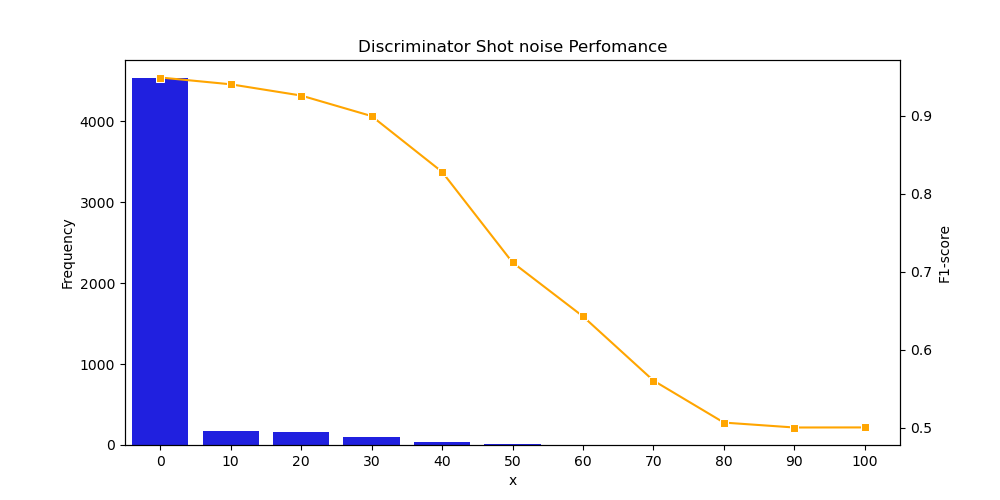}
    \caption{Shot}
\end{subfigure}
\begin{subfigure}[t]{0.3\textwidth}
    \centering
    \includegraphics[width=\linewidth]{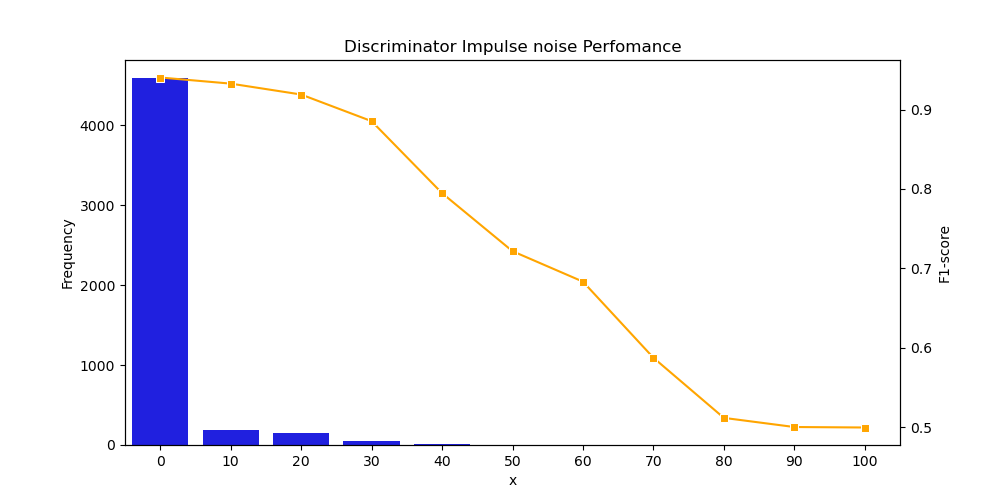}
    \caption{Impulse}
\end{subfigure}

\vspace{0.25cm}

% -------- Row 2: Blur --------
\begin{subfigure}[t]{0.3\textwidth}
    \centering
    \includegraphics[width=\linewidth]{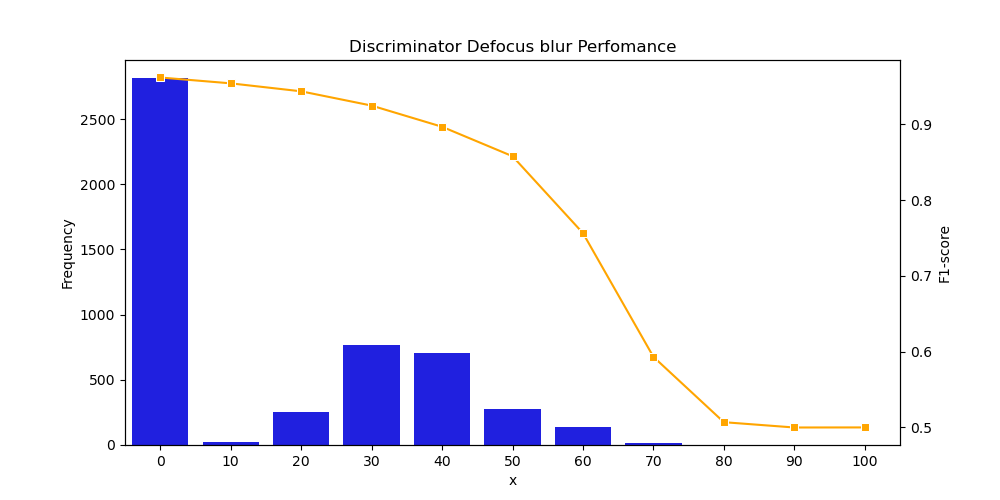}
    \caption{Defocus}
\end{subfigure}
\begin{subfigure}[t]{0.3\textwidth}
    \centering
    \includegraphics[width=\linewidth]{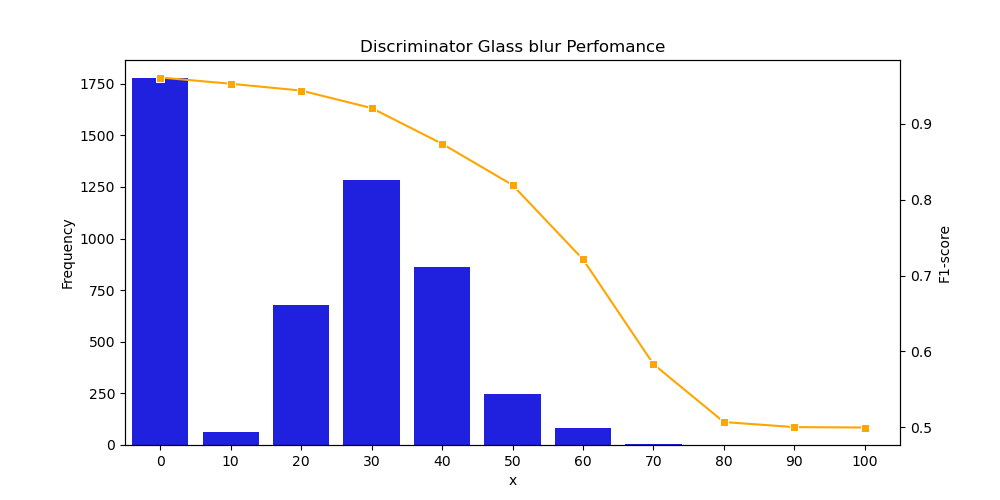}
    \caption{Glass}
\end{subfigure}
\begin{subfigure}[t]{0.3\textwidth}
    \centering
    \includegraphics[width=\linewidth]{images/eval_motion_blur.png}
    \caption{Motion}
\end{subfigure}

\vspace{0.25cm}

% -------- Row 3: Blur (cont.) --------
\begin{subfigure}[t]{0.3\textwidth}
    \centering
    \includegraphics[width=\linewidth]{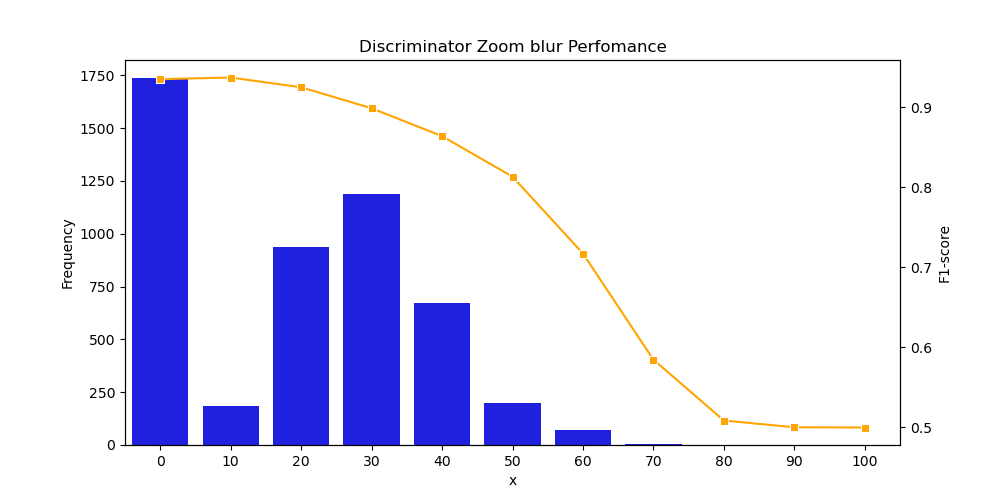}
    \caption{Zoom}
\end{subfigure}
\begin{subfigure}[t]{0.3\textwidth}
    \centering
    \includegraphics[width=\linewidth]{images/eval_fog.png}
    \caption{Fog}
\end{subfigure}
\begin{subfigure}[t]{0.3\textwidth}
    \centering
    \includegraphics[width=\linewidth]{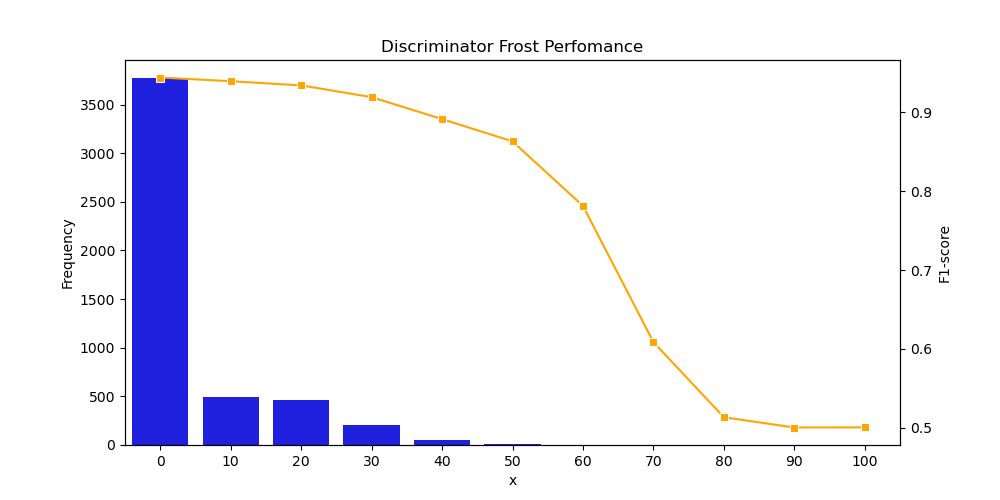}
    \caption{Frost}
\end{subfigure}

\vspace{0.25cm}

% -------- Row 4: Weather --------
\begin{subfigure}[t]{0.3\textwidth}
    \centering
    \includegraphics[width=\linewidth]{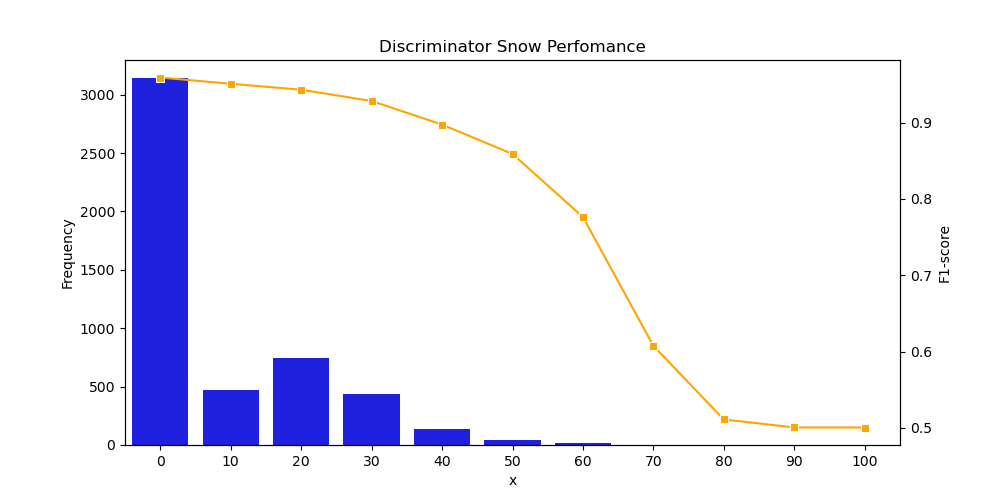}
    \caption{Snow}
\end{subfigure}
\begin{subfigure}[t]{0.3\textwidth}
    \centering
    \includegraphics[width=\linewidth]{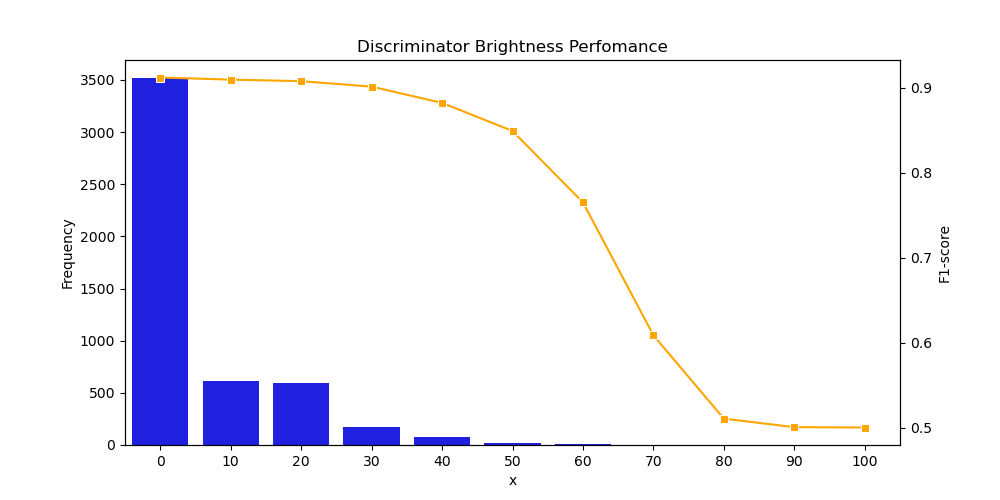}
    \caption{Brightness}
\end{subfigure}
\begin{subfigure}[t]{0.3\textwidth}
    \centering
    \includegraphics[width=\linewidth]{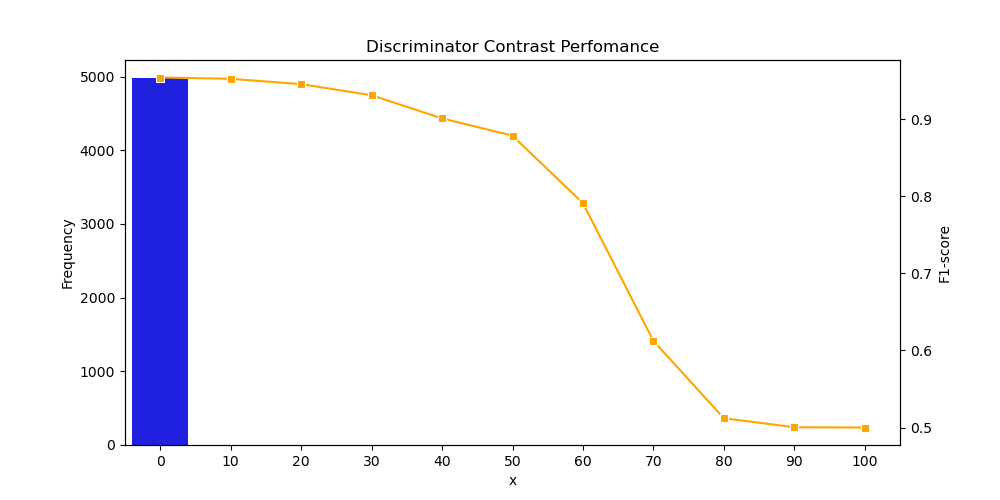}
    \caption{Contrast}
\end{subfigure}

\vspace{0.25cm}

% -------- Row 5: Digital --------
\begin{subfigure}[t]{0.3\textwidth}
    \centering
    \includegraphics[width=\linewidth]{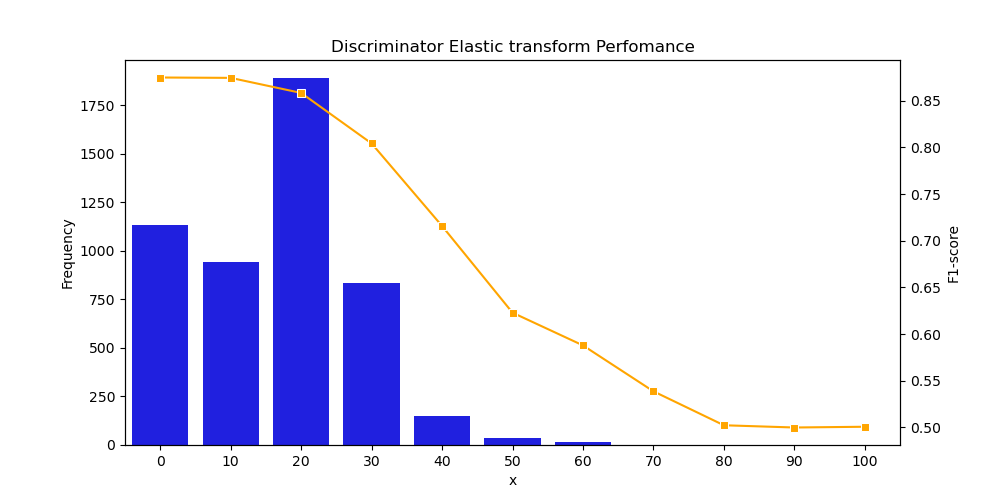}
    \caption{Elastic}
\end{subfigure}
\begin{subfigure}[t]{0.3\textwidth}
    \centering
    \includegraphics[width=\linewidth]{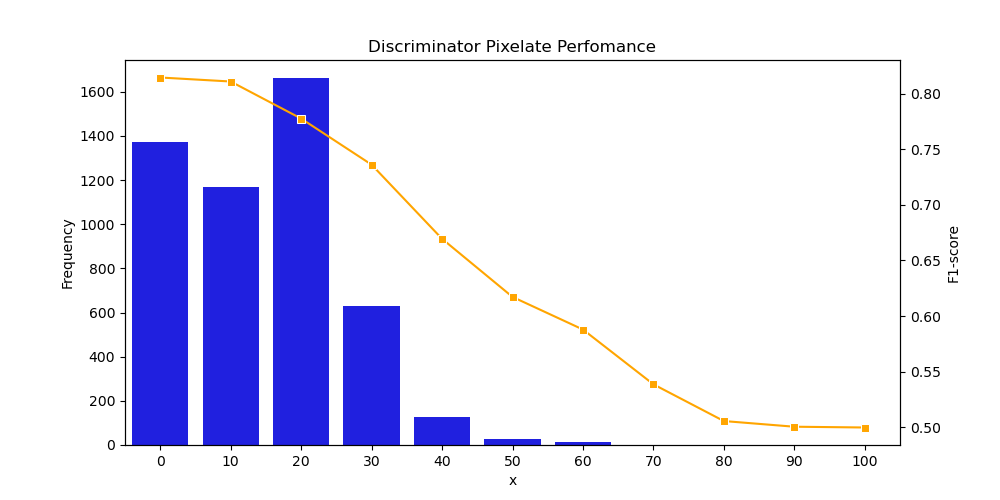}
    \caption{Pixelate}
\end{subfigure}
\begin{subfigure}[t]{0.3\textwidth}
    \centering
    \includegraphics[width=\linewidth]{images/eval_jpeg_compression.png}
    \caption{JPEG}
\end{subfigure}

\caption{Discriminator-guided adaptive stopping behavior across all 15 corruption types. Each subfigure shows the distribution of adaptive stopping timesteps $t^*$ (blue histogram) together with discriminator performance across forward diffusion steps (orange curve). Different corruption types exhibit distinct stopping behaviors, reflecting varying amounts of diffusion required to suppress domain-specific cues.}
\label{ind_perf_all}
\end{figure*}

\section{Adaptive Stopping Behavior of the Discriminator}

Fig.~\ref{ind_perf_all} provides a comprehensive view of the discriminator-guided adaptive stopping behavior across all 15 corruption types. Each subplot reports the distribution of adaptive stopping timesteps $t^*$ together with discriminator performance as a function of the forward diffusion depth. Across corruptions, discriminator confidence consistently decreases as diffusion progresses, indicating the gradual suppression of domain-specific cues. Notably, the rate of this degradation varies substantially across corruption types.

Noise-based corruptions, such as Gaussian, Shot, and Impulse noise, typically require deeper diffusion before discriminator confidence approaches chance level, leading to stopping distributions concentrated at larger $t^*$. In contrast, spatially structured corruptions, including blur and weather effects, exhibit earlier stopping behavior, suggesting that fewer diffusion steps are sufficient to obscure corruption-specific cues. Digital corruptions show intermediate behavior, with stopping distributions spanning moderate diffusion depths.

These observations confirm that the adaptive stopping mechanism responds to corruption-dependent characteristics rather than following a fixed diffusion schedule. By selecting the stopping timestep on a per-sample basis, the method avoids excessive diffusion when limited perturbation is sufficient, while enabling deeper diffusion when required. This behavior provides additional insight into the balanced robustness observed across corruption families in Tab.~1 of the main paper. \\

% T-SNE CORRUPTION DOMAIN SAMPLES LATENTS
%\noindent

\begin{figure}[ht!]
    \centering

    \begin{subfigure}[t]{0.48\linewidth}
        \centering
        \includegraphics[width=\linewidth]{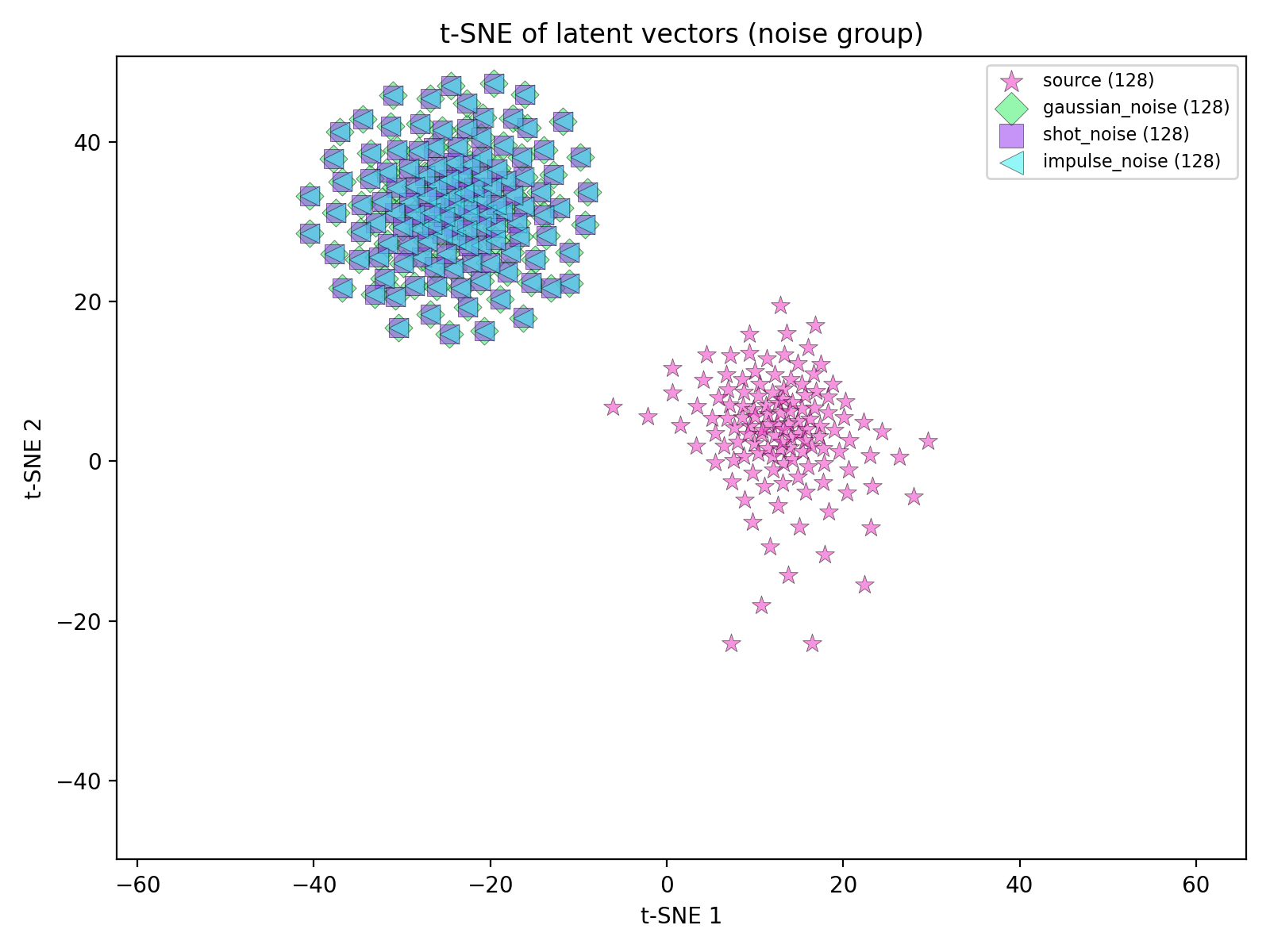}
        \caption{Noise corruptions}
        \label{fig:tsne_noise}
    \end{subfigure}
    \hfill
    \begin{subfigure}[t]{0.48\linewidth}
        \centering
        \includegraphics[width=\linewidth]{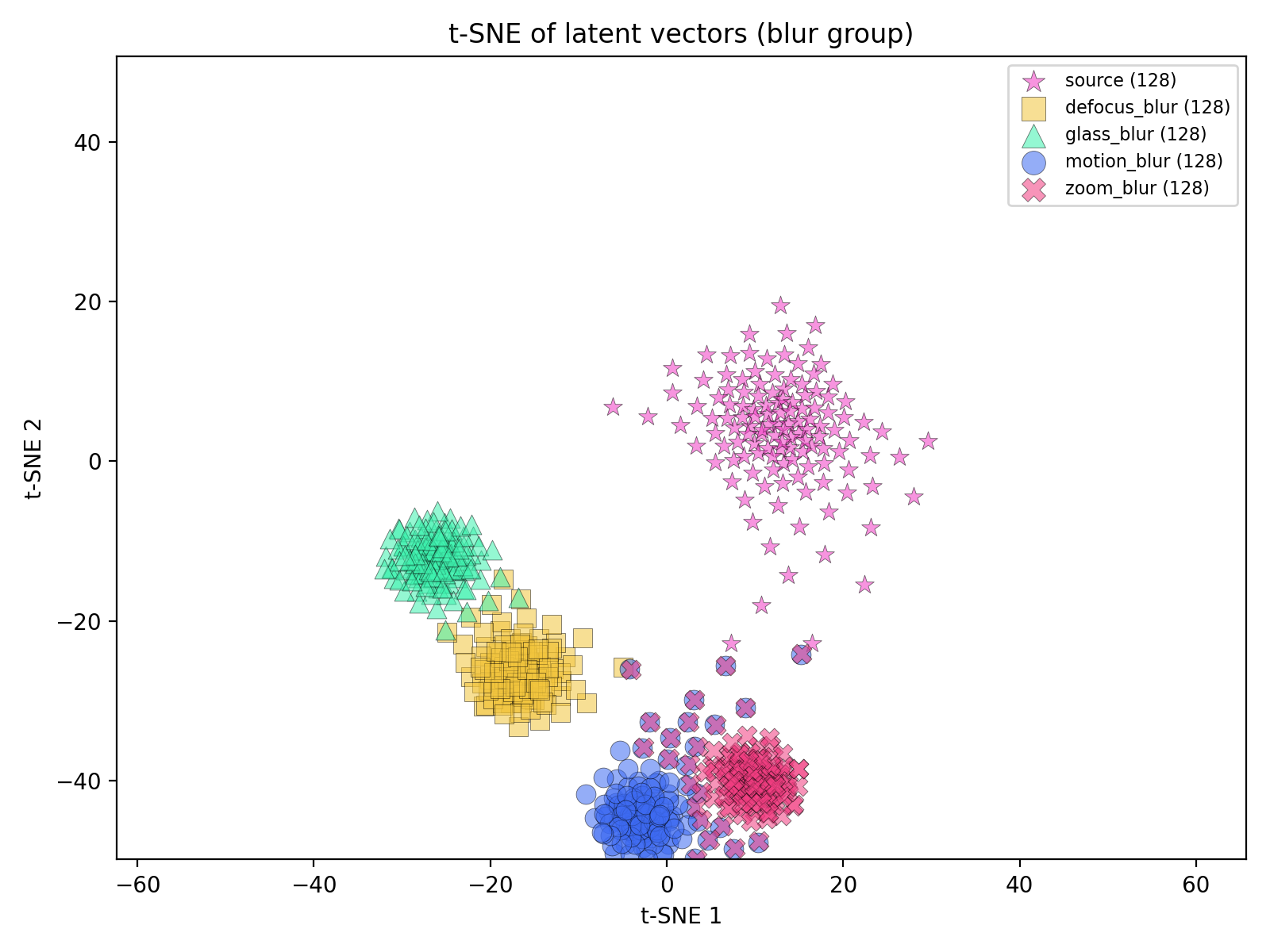}
        \caption{Blur corruptions}
        \label{fig:tsne_blur}
    \end{subfigure}

    \vspace{0.4cm}

    \begin{subfigure}[t]{0.48\linewidth}
        \centering
        \includegraphics[width=\linewidth]{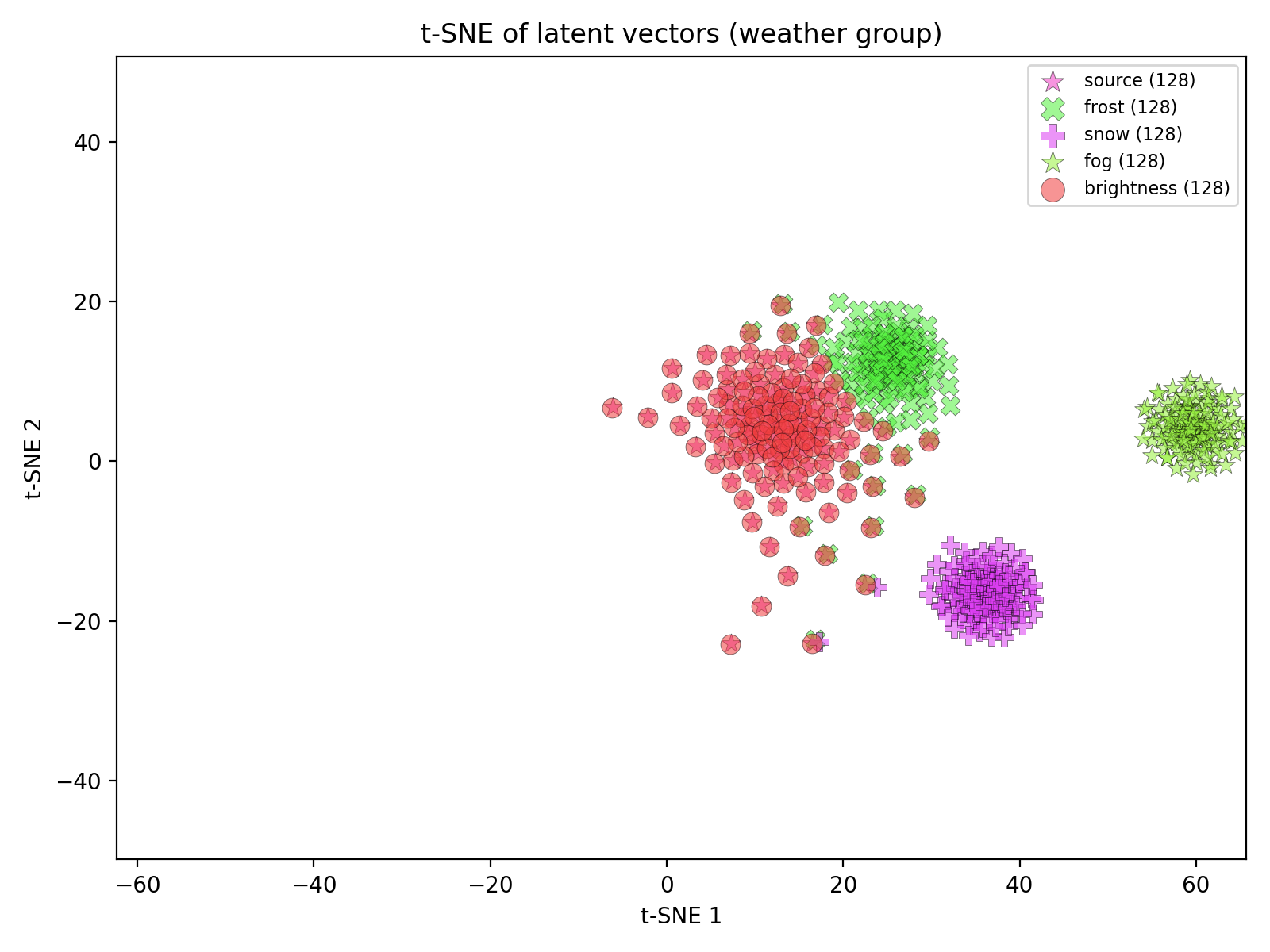}
        \caption{Weather corruptions}
        \label{fig:tsne_weather}
    \end{subfigure}
    \hfill
    \begin{subfigure}[t]{0.48\linewidth}
        \centering
        \includegraphics[width=\linewidth]{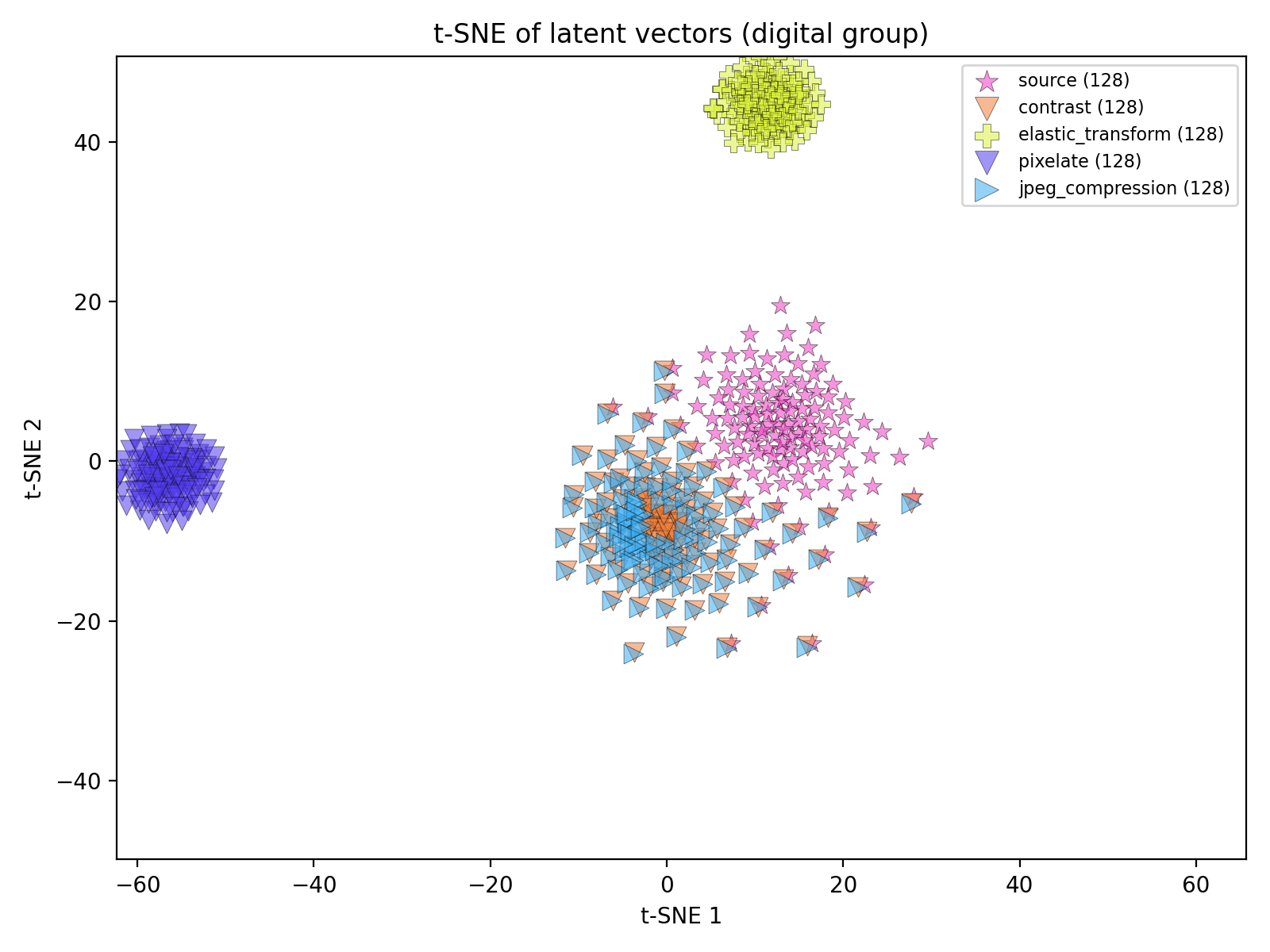}
        \caption{Digital corruptions}
        \label{fig:tsne_digital}
    \end{subfigure}

    \caption{t-SNE visualization of feature representations for different corruption families and the source domain. Each subfigure shows the relative clustering of corrupted samples and source samples, providing an estimate of their similarity in feature space. Noise-based corruptions form more distinct clusters, while blur, weather, and digital corruptions exhibit closer alignment with the source domain.}
    \label{fig:tsne_all}
\end{figure}

\section{Qualitative Visualization of Feature Representations}

Fig. \ref{fig:tsne_all} presents t-SNE visualizations of feature representations for different corruption families alongside the source domain. These plots are intended solely to provide qualitative intuition regarding how corrupted samples from different families relate to the source distribution in feature space. Consistent with the quantitative analyses in the main paper, noise-based corruptions tend to form more separated clusters, while blur, weather, and digital corruptions exhibit closer alignment with source samples. We emphasize that t-SNE embeddings are sensitive to hyperparameters and do not preserve global distances; therefore, these visualizations are not used to motivate design choices or draw quantitative conclusions, but rather to offer additional qualitative context.

\bibliographystyle{plain}
\bibliography{references} % The name of your .bib file
\end{document}